\documentclass[sigconf,nonacm]{acmart}
\usepackage{booktabs,amsmath,enumitem,graphicx,subcaption}
\begin{document}
\title{Off-Policy Evaluation for Semantic ID Recommenders:\\ Does the Model's Own Code Hierarchy Help?}

\author{Artem Betlei}
\email{a.betlei@criteo.com}
\affiliation{%
  \institution{Criteo AI Lab}
  \city{Grenoble}
  \country{France}
}

\renewcommand{\shortauthors}{Betlei et al.}

\begin{abstract}
Generative recommenders increasingly emit \emph{semantic IDs} (SIDs): each item is a short sequence of
hierarchical discrete codes from a residual quantizer, decoded autoregressively. Before spending scarce
A/B-test, a team may decide \emph{offline} which decoder or reranking variants are worth testing - a
job for off-policy evaluation (OPE). We ask a simple question: can the model's \emph{own} SID tree serve
as the action abstraction for that OPE? Our answer has three parts. (i)~Under the near-argmax logging real
recommenders use, per-item OPE is hopeless - as item-level effective sample size is usually small on
production logs - but marginalizing items to code-prefix \emph{clusters} restores estimable support and cuts error. (ii)~This gain is thanks to \emph{coarsening}, not to the hierarchy specifically; but the
SID tree is what makes coarsening \emph{feasible} in a generative system - each
cluster's mass is exactly and cheaply returned by the decoder, whereas flat clustering requires enumerating item/leaf masses that a
code-only decoder does not directly expose. (iii)~Resolution depth is the operative knob - coarser under
scarce support - and a conditional
bias bound links the coarsening bias to the quantizer's \emph{worst-case} reconstruction residual and the
target - logging divergence.
\end{abstract}
\maketitle

\section{Introduction}
Generative retrieval - encoding each item as a short sequence of hierarchical
discrete codes called \emph{semantic ID} (SID), and decoding next item SID autoregressively - is moving into production
recommenders~\cite{rajput2023tiger,wang2024letter}. Every change to such a system (e.g. new decoder, decoding
temperature, or reranking rule) shifts the distribution of items it surfaces. Deciding which candidate
changes deserve scarce, costly A/B-test is the role of \emph{off-policy evaluation} (OPE): estimate
a candidate policy's online reward from logs collected under the incumbent, so candidates can be ranked and
filtered before live test.

The obstacle is how production systems log. Exploration is typically \emph{near-argmax} - weighted by item
score, not uniform - so a handful of items dominate each impression and the importance weights of per-item
inverse propensity scoring (IPS) explode. On our production logs the item-level effective sample size (ESS)
is only ${\approx}0.003$ of the raw count: per-item OPE is not a usable decision tool. Our scope is the
\emph{item-selection} policy on a fixed catalog (logging and target choose items from the same pool); we do
not address a changed candidate generator, which alters the action set and may surface items with no logged
support~\cite{wang2025twostage}.

A SID model hands us a ready-made action abstraction: its code prefixes form a nested partition
of the catalog. Marginalizing per-item propensities up to a code-prefix \emph{cluster} pools support and
makes weights estimable again, at the cost of a within-cluster bias. Crucially, an autoregressive decoder
already produces each prefix's probability \emph{exactly}, as a product of per-level conditionals, without
summing over the catalog - so code depth is a native, cheap resolution knob. Whether a SID tree is
\emph{reward-coherent} enough for this to pay off is not obvious, and depends on the codes.

We provide following contributions: i) We frame OPE where the action abstraction \emph{is} the model's SID tree, and give a
production diagnostic showing per-item OPE is infeasible under realistic near-argmax logging. ii) We propose then to marginalize to code prefixes; a conditional bias bound
(Prop.~\ref{prop:bias}) ties the coarsening bias of cluster marginal-IPS to the quantizer's worst-case
reconstruction residual and the target - logging divergence. 
iii) We conduct series of experiments including support recovery, distinct values of SID tree, cross-domain comparison of the estimators, embedding sources, quantizers and depth selection from logged data.

\paragraph{Related work.} Large-action OPE supplies the estimators we reuse at code granularity: \emph{MIPS}
reweights over a coarser action embedding, trading variance for marginalization bias~\cite{saito2022mips};
\emph{OffCEM} adds a reward model plus a residual weight, so the bias is only what the model
misses~\cite{saito2023offcem}; \emph{Policy Convolution} (PC) smooths over an embedding kernel whose
bandwidth plays the role of resolution~\cite{sachdeva2023pc}; \emph{POTEC} decomposes into clusters for
policy \emph{learning}~\cite{peng2024potec}. All take a flat embedding or ad-hoc clustering as given. The
nearest generative-retrieval work adapts doubly-robust estimation \emph{inside} an off-policy REINFORCE
learning objective over dense two-tower embeddings, not over a code
tree~\cite{matveev2026offpolicygenret}; to our knowledge no prior work evaluates policies on the SID tree
itself. Others \emph{learn} action embeddings~\cite{cief2024actionemb} or select estimators and hyperparameters from
data~\cite{udagawa2023selection,felicioni2024autoope} - building blocks for the open depth-selection step
(RQ5). Closest in spirit, the marginal density-ratio estimator~\cite{taufiq2023mdr} learns its ratio
\emph{from} the base per-action propensities, so it reintroduces the item-level dependence a code-only
decoder does not expose. We instead test the retriever's \emph{native} tree.

\section{OPE on Semantic ID Trees}\label{sec:sidope}
\emph{Setup and notation.} A logging policy $\pi_0(a\mid x)$ generates logs $\{(x_i,a_i,r_i)\}_{i=1}^M$:
context $x_i$, chosen item $a_i$ from catalog $\mathcal{A}$, reward $r_i$ with mean $q(x,a)$. We want the
value of a target policy, $V(\pi_e)=\mathbb{E}_x\,\mathbb{E}_{a\sim\pi_e}[q(x,a)]$. Each item $a$ has a SID
$c(a)=(c_1,\dots,c_L)$ from a residual quantizer of an item embedding $z(a)$; the length-$\ell$ prefix
$c_{\le\ell}(a)$ names a \emph{cluster} $A_c=\{a':c_{\le\ell}(a')=c\}$. A policy's mass on that cluster is
$\pi(c\mid x)=\sum_{a\in A_c}\pi(a\mid x)$. Item-level IPS reweights by $\pi_e(a)/\pi_0(a)$ and needs
per-item overlap; the self-normalized variant (SNIPS) divides by the summed weights to curb variance.

\emph{Marginalizing to the tree.} The core move is to reweight at cluster granularity instead:
$\hat V_\ell=\frac1M\sum_i \tfrac{\pi_e(c_{\le\ell}(a_i)\mid x_i)}{\pi_0(c_{\le\ell}(a_i)\mid x_i)}\,r_i$,
a hierarchical MIPS estimator~\cite{saito2022mips}. It corrects the target's distribution \emph{over}
clusters while leaving the logging distribution \emph{within} each cluster untouched, so its only bias is
within-cluster target - logging mismatch (Section~\ref{sec:bias}). It requires only \emph{cluster
positivity} - $\pi_0(c\mid x)>0$ wherever $\pi_e(c\mid x)>0$ - far weaker than per-item positivity. Coarser
$\ell$ increases support but raises the within-cluster bias: depth is the bias - variance knob.

\emph{Why the tree, not any clustering.} For an autoregressive SID policy the cluster mass is not a sum
over $\mathcal{A}$ but the exact prefix probability
$\pi(c_{\le\ell}\mid x)=\prod_{t\le\ell}\pi(\mathrm{code}_t\mid x,\mathrm{code}_{<t})$, available in $O(\ell)$
directly from the decoder. Forming the same mass for a \emph{flat} clustering would require the item-level
propensities $\pi_0(a\mid x)$ and an $O(|\mathcal{A}|)$ sum - and a generative retriever emits codes, not a
normalized item distribution, so those item propensities are typically unavailable. The SID tree is thus the
cluster representation one can actually run online. (If several items share a full code, that leaf is simply
a small cluster; nothing changes.) We verify this equivalence is exact, not approximate: building an
autoregressive tree policy and reading masses as prefix products reproduces the item-summed cluster
marginals to machine precision at every level (App.~\ref{app:data}) - so the item-sum cluster masses used in
our controlled experiments are precisely what a decoder returns for free.

\section{Bias and Quantization Quality}\label{sec:bias}
\emph{Intuition.} Because marginalization exactly matches the target's cluster masses, the estimator
can only err \emph{inside} clusters: its bias is how much $\pi_e$ and $\pi_0$ disagree within a cluster,
weighted by how much reward varies there - the informal counterpart of OffCEM's within-cluster residual
effect~\cite{saito2023offcem}. Reward-coherent codes make that variation small. Quantization quality is the
natural proxy: items in a cluster share a reconstruction $\hat z_\ell$, so if reward is smooth in the
embedding, within-cluster reward spread is controlled by the reconstruction residual.

\emph{Bound.} Let $\hat z_\ell(a)$ reconstruct $z(a)$ from its first $\ell$ codes, let
$\delta_\ell(a)=\|z(a)-\hat z_\ell(a)\|$, and let $\delta^{\max}_\ell=\max_a\delta_\ell(a)$ be the
worst-case residual; $\varepsilon_\ell=\mathbb{E}_{a\sim\mu}\,\delta_\ell(a)^2$ is the mean-squared
reconstruction error a quantizer is trained to minimize. Write
$\mathrm{TV}^{\mathrm{in}}_\ell(x)=\sum_c\pi_e(c\mid x)\,
\mathrm{TV}\big(\tilde\pi_e(\cdot\mid c,x),\tilde\pi_0(\cdot\mid c,x)\big)$ for the $\pi_e$-weighted
\emph{within-cluster} disagreement, where $\tilde\pi(\cdot\mid c,x)$ is the conditional a policy induces
inside cluster $A_c$. Assume \textbf{(A0)} cluster positivity, known propensities and $\mathbb{E}[r\mid x,a]=q(x,a)$;
\textbf{(A1)} \emph{no direct effect}, $q(x,a)=\tilde q(x,z(a))$ - MIPS's assumption~\cite{saito2022mips},
imposed here at the \emph{full} embedding so coarsening bias is bounded rather than zero; \textbf{(A2)}
$\tilde q(x,\cdot)$ is $L_q$-Lipschitz on a set containing every item embedding and cluster reconstruction.
\begin{proposition}\label{prop:bias}
Let $\mathrm{Bias}(\hat V_\ell)=\mathbb{E}[\hat V_\ell]-V(\pi_e)$ be the error of the level-$\ell$ cluster
marginal-IPS estimator ($\mathrm{MIPS}_{\mathrm{hier}}$) against the true target value. Under (A0) - (A2),
\begin{align*}
\big|\mathrm{Bias}(\hat V_\ell)\big|
&\;\le\;2L_q\,\delta^{\max}_\ell\;\mathbb{E}_x\big[\mathrm{TV}^{\mathrm{in}}_\ell(x)\big] \le\;4L_q\,\delta^{\max}_\ell\;\mathbb{E}_x\big[\mathrm{TV}(\pi_e,\pi_0)\big],
\end{align*}
and the constant in the second, item-level form cannot be improved (proof in App.~\ref{app:proof}). If in
addition $\delta^{\max}_\ell\le\kappa\sqrt{\varepsilon_\ell}$, both forms hold with $\delta^{\max}_\ell$
replaced by $\kappa\sqrt{\varepsilon_\ell}$.
\end{proposition}
\emph{Reading the bound.} Two quantities control the bias, and mean reconstruction error is neither.
The \emph{worst-case} residual comes first: a finer codebook tightens the bound only insofar as it improves
the worst-reconstructed items, so it can lower $\varepsilon_\ell$ while leaving $\delta^{\max}_\ell$
untouched - and reaching the $\varepsilon_\ell$ form costs a codebook-specific $\kappa$, so
$\varepsilon_\ell$ is not comparable across tokenizers or sources. The second is \emph{within-cluster}
disagreement: coarsening creates no bias where policies agree conditionally, however far apart they are
across clusters. The item-level form discards that, hence the ordering above.

Three limits, detailed in App.~\ref{app:proof}: the bound covers raw cluster marginal-IPS only; it beats the
trivial $|\mathrm{Bias}|\le1$ only while the quantity
$2L_q\delta^{\max}_\ell\mathbb{E}_x[\mathrm{TV}^{\mathrm{in}}_\ell]$ stays below $1$;
and (A1) holds only approximately in our testbeds, so we use it to reason about the coarsening
\emph{mechanism}, not to predict measured bias. \emph{Variance} moves the other way: cluster marginal-IPS has
the usual marginalized-weight second moment~\cite{saito2022mips}, which falls as coarsening pools support.
Closed-form variance is unwieldy for the estimators we headline, so we track it via ESS (an inverse-variance
proxy; RQ1's collapse is a weight-variance blow-up) and bootstrap CIs.

\section{Experiments}
We answer six questions. \textbf{RQ1}: is per-item OPE feasible on real reco logs? \textbf{RQ2}: does
code-cluster marginalization recover support, and how do known estimators compare at code granularity?
\textbf{RQ3}: \emph{when} does the hierarchy help, and how does the best resolution move with support?
\textbf{RQ4}: how do code source and tokenizer affect it? \textbf{RQ5}: can adaptive resolution beat a fixed
level? \textbf{RQ6}: does it generalize to a second dataset?

\subsection{Datasets and protocol}\label{sec:proto}
\textbf{Production logs} (motivation, RQ1). A large e-commerce recommender explores by score-weighted
(Plackett - Luce) sampling, so served-item propensities are recoverable from the logged normalizer. Lacking a
randomized-exposure oracle, we use these logs for support diagnostics only (App.~\ref{app:data}).

\textbf{KuaiRand} (\emph{primary} controlled testbed, RQ2 - RQ5). Its random-exposure
slice~\cite{gao2022kuairand} shows items uniformly at random, so per-item click rates are unbiased and the
oracle is identified without exposure confounding. We split by row: one half sets the click rates building
the logging and target policies, the disjoint half is the evaluation truth $q_{\mathrm{eval}}$ - both the
data-generating process (DGP) for rewards $r\sim\mathrm{Bernoulli}(q_{\mathrm{eval}}(a))$ and the oracle
value $V(\pi_e)=\sum_a\pi_e(a)\,q_{\mathrm{eval}}(a)=0.499$. No estimator sees evaluation outcomes through
the policies: the protocol is \emph{leakage-free} (App.~\ref{app:data}). It is a catalog bandit over
$|\mathcal{A}|{=}7339$ items with content-metadata, semantic, collaborative and fused code trees
(\S\ref{sec:rq4}).

\textbf{Amazon Reviews} (cross-domain replication, RQ6). Musical Instruments from Amazon
Reviews~2023~\cite{hou2024amazon}: $|\mathcal{A}|{=}21009$ items with ${\ge}10$ reviews per held-out half,
same reward template with $q(a)=P(\text{5-star}\mid a)$ as oracle and DGP. Ratings are self-selected rather
than randomly exposed, so that oracle is confounded and Amazon \emph{replicates} the pattern rather than
identifying it.

\textbf{Estimators.} All are existing methods, run at two granularities marked by a subscript
($\mathrm{EST}_{\mathrm{item}}$ on items, $\mathrm{EST}_{\mathrm{hier}}$ on SID prefix clusters):
item-level IPS and SNIPS; cluster IPS and SNIPS ($\mathrm{MIPS}_{\mathrm{hier}}$,
$\mathrm{SNIPS}_{\mathrm{hier}}$), reweighting by $\pi_e(c)/\pi_0(c)$; the cluster direct method
$\mathrm{DM}_{\mathrm{hier}}$; OffCEM~\cite{saito2023offcem}, whose original configuration pairs the
cluster-level residual weight with an \emph{action}-level reward model - a standard doubly-robust
form~\cite{dudik2011dr} - reported as $\mathrm{OffCEM}^{\mathrm{act}}_{\mathrm{hier}}$, alongside a
coarsened cluster-level variant $\mathrm{OffCEM}_{\mathrm{hier}}$ we add for symmetry; and
PC~\cite{sachdeva2023pc}, an embedding kernel whose bandwidth is a continuous resolution knob (so no
$\mathrm{hier}$ variant), at untuned $h$. App.~\ref{app:data} gives formulas and the two deliberate
departures from published OffCEM.

A \emph{seed} is one simulated log ($M$ draws from $\pi_0$ plus simulated rewards); RMSE is taken over seeds,
$200$ for the headline comparison and support sweep, $100$ per fit across $10$ fits for the quantizer
ablation, $80$ for Amazon. We report mean $\pm$ the half-width of a $95\%$ bootstrap percentile interval and
call a difference \emph{significant} when its \emph{paired} interval excludes zero, with no multiplicity
correction. 

\textbf{Default tree.} Codes are residual K-Means unless stated; RQ4 finds the tokenizer does not move these results.

\textbf{Depth selection.} Unless stated otherwise $\mathrm{hier}$ estimators pick the level by
error against the \emph{oracle} on a disjoint validation split of seeds, so they are \emph{oracle-selected
diagnostics} - an upper bound on what depth selection can buy. Logged-data-only, deployable alternative
(SLOPE~\cite{su2020slope}, estimator variance alone) is reported (RQ3).

\subsection{Per-item OPE is infeasible (RQ1)}
RQ1 asks whether per-item OPE is usable on real recommender logs. It is not, which motivates everything that
follows. 
On production logs the served item's propensity is highly concentrated (median $0.04-0.15$; $6-16\%$
of impressions above $0.9$), so the item-level effective sample size (ESS) is only ${\approx}0.003$ of the sample
(Table~\ref{tab:tierAB}, App.~\ref{app:data}). So per-item IPS is not a credible tool, motivating for
coarsening.

\subsection{Coarsening recovers support and improves accuracy (RQ2)}
RQ2 has two parts: does marginalizing to code clusters restore the support item-level OPE lacks, and 
does the estimator choice matter?

\textbf{Support is restored, and converts into accuracy.} On the candidate-logged production slice,
coarsening lifts the greedy-target ESS by ${\sim}1.8\times$ (Table~\ref{tab:tierAB}). On the KuaiRand
oracle, where error is measurable, Table~\ref{tab:kuairand} compares both granularities: $\mathrm{SNIPS}_{\mathrm{item}}$
attains $0.155$ RMSE against $0.087$ for $\mathrm{SNIPS}_{\mathrm{hier}}$.

\textbf{At code granularity, no estimator dominates.} Hierarchical MIPS/SNIPS, cluster DM, and both
OffCEM variants perform close to one another. 
PC, which uses no code tree, beats the best cluster estimator by a paired $-0.009$ at $h{=}0.1$, but that advantage
is bandwidth-selected - PC degrades at $h{=}0.2$ - and it
needs the item-level propensities of \S\ref{sec:sidope}, so a code-only deployment cannot count on it.
What generalizes across the estimators is change in granularity.

\begin{table}[ht]\centering\small
\caption{RQ2, KuaiRand, leakage-free ($|\mathcal{A}|{=}7339$, $M{=}5000$; $\mathrm{hier}$ at the
oracle-selected level). (a) Estimator comparison. (b) SID tree vs.\ flat $k$-means at matched cluster counts (metadata,
$\mathrm{OffCEM}^{\mathrm{act}}_{\mathrm{hier}}$), each row refitted under $10$ shared clustering seeds;
$\Delta$ is the paired tree${-}$flat difference.}
\begin{subtable}[t]{0.5\columnwidth}\centering
\resizebox{0.8\linewidth}{!}{%
\begin{tabular}{lc}\toprule
estimator & RMSE\\\midrule
$\mathrm{IPS}_{\mathrm{item}}$    & $1.57\pm0.7$\\
$\mathrm{SNIPS}_{\mathrm{item}}$  & $0.155\pm0.022$\\
$\mathrm{MIPS}_{\mathrm{hier}}$   & $0.094\pm0.009$\\
$\mathrm{SNIPS}_{\mathrm{hier}}$  & $0.087\pm0.007$\\
$\mathrm{DM}_{\mathrm{hier}}$     & $0.087\pm0.007$\\
$\mathrm{OffCEM}_{\mathrm{hier}}$ & $0.088\pm0.007$\\
$\mathrm{OffCEM}^{\mathrm{act}}_{\mathrm{hier}}$ & $0.086\pm0.007$\\
PC ($h{=}0.1$)           & $\mathbf{0.076}\pm0.006$\\
PC ($h{=}0.2$)           & $0.123\pm0.005$\\\bottomrule
\end{tabular}}
\caption{}\label{tab:kuairand}
\end{subtable}\hfill
\begin{subtable}[t]{0.5\columnwidth}\centering
\resizebox{0.8\linewidth}{!}{%
\begin{tabular}{ccc}\toprule
\# clust.\ (lvl) & tree & flat\\\midrule
8 (L1)   & 0.112 & 0.114\\
40 (L2)  & 0.097 & 0.097\\
126 (L3) & 0.094 & 0.093\\
288 (L4) & 0.099 & 0.090\\\bottomrule
\end{tabular}}
\caption{}\label{tab:treeflat}
\end{subtable}
\end{table}

\emph{Is the gain due to the tree, or to coarsening in general?} To separate the two we replace the SID
tree with a flat $k$-means clustering of the \emph{same} embedding at matched cluster counts (Table~\ref{tab:treeflat}). Flat
clustering recovers the item$\to$cluster gain in full, and at no level is either representation
significantly more accurate than the other: all four paired differences contain zero. Support recovery is
therefore a property of coarsening, not of the hierarchy - and the two representations are, on accuracy,
interchangeable.

The tree's consistent advantage is instead feasibility. A deployed generative retriever reads cluster masses
off its decoder exactly, in $O(\ell)$, whereas a flat clustering of the same catalog needs the item-level
propensities such a system does not expose (\S2). That is a separate axis from the RMSE comparison above.

\emph{Ranking candidates, not just estimating values.} Gating is a ranking problem: a team must order
candidates, not measure any one precisely. We rank $16$ simulated variants against the oracle (Table~\ref{tab:select}).

The method applies to changes sharing a single SID tree - a new decoder, decoding temperature, or reranking
rule - since both policies' masses come from the same partition; a re-tokenization would move the tree
itself and is out of scope. Within that scope we test a \emph{cross-cluster} family, whose variants shift
preference along embedding directions and so move mass between prefixes. Here
$\mathrm{SNIPS}_{\mathrm{hier}}$ ranks better than $\mathrm{SNIPS}_{\mathrm{item}}$, with the margin
widening as support grows scarcer (Kendall $\tau$ $0.13$ vs.\ $0.08$ at $M{=}1000$).

Two boundaries apply. When variants differ only \emph{within} clusters, coarsening averages away the very
differences that distinguish them, and $\mathrm{SNIPS}_{\mathrm{item}}$ ranks at least as well. And depth here is pre-specified at
level 3 rather than validated, while the deployment-regret gap falls within confidence intervals. We
therefore claim improved rank correlation on cross-cluster changes, a narrower result than a general
selection guarantee.

\begin{table}[ht]\centering\small
\caption{RQ2 (selection). Ranking $16$ candidate variants off one log against the oracle ranking (level 3,
pre-specified). For realistic
\emph{cross-cluster} variants, $\mathrm{SNIPS}_{\mathrm{hier}}$ ranks better than
$\mathrm{SNIPS}_{\mathrm{item}}$, more so under scarce support.}
\label{tab:select}
\resizebox{0.6\linewidth}{!}{%
\begin{tabular}{llcc}\toprule
$M$ & estimator & Kendall $\tau$ & top-3 hit\\\midrule
$1000$ & $\mathrm{SNIPS}_{\mathrm{item}}$ & $0.08\pm0.03$ & $10\%$\\
$1000$ & $\mathrm{SNIPS}_{\mathrm{hier}}$ & $\mathbf{0.13}\pm0.03$ & $12\%$\\
\midrule
$5000$ & $\mathrm{SNIPS}_{\mathrm{item}}$ & $0.15\pm0.03$ & $6\%$\\
$5000$ & $\mathrm{SNIPS}_{\mathrm{hier}}$ & $\mathbf{0.18}\pm0.02$ & $\mathbf{14\%}$\\\bottomrule
\end{tabular}
}
\end{table}

\subsection{When does the hierarchy help? (RQ3)}
RQ2 established that coarsening helps on average; RQ3 asks when, and how far to coarsen. Available support
governs both.

\begin{figure}[ht]\centering
\includegraphics[width=\columnwidth]{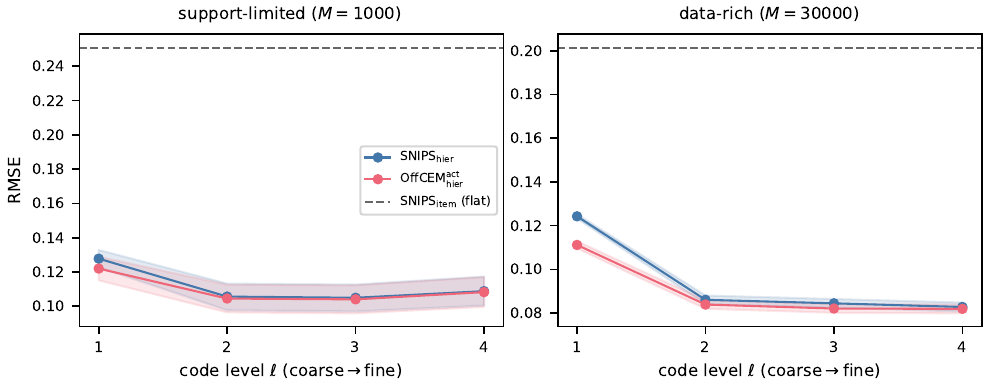}
\caption{RQ3. KuaiRand RMSE by code level (coarse$\to$fine), metadata source, two support regimes
($200$ seeds, $95\%$ bootstrap bands). Under scarce support (left) an
intermediate level is best; with more data (right) the optimum moves finer.}
\label{fig:depth}
\end{figure}

Fig.~\ref{fig:depth} varies the code level directly. Under scarce support, code clustering halves $\mathrm{SNIPS}_{\mathrm{item}}$ RMSE and an intermediate level minimizes error - coarse enough to pool support,
fine enough to keep clusters reward-coherent; optimum moves finer with more data. 


Table~\ref{tab:sweeps} varies support directly. Raising $M$ from $1000$ to $10000$ narrows the granularity
gap: coarsening buys most where support is scarcest, which is what the bias - variance account predicts.

\begin{table}[ht]\centering\small
\caption{RQ3. KuaiRand support sweep, leakage-free, content-metadata codes ($|\mathcal{A}|{=}7339$, $200$
seeds; RMSE). \emph{oracle} is $\mathrm{SNIPS}_{\mathrm{hier}}$ at the oracle-selected level (an upper
bound); \emph{SLOPE} selects that level from logged data alone - applied to
$\mathrm{MIPS}_{\mathrm{hier}}$, whose per-level standard errors the rule needs - and is therefore deployable. SLOPE beats $\mathrm{SNIPS}_{\mathrm{item}}$ at every $M$, capturing $67$ - $81\%$ of the available gain. The $M{=}5000$ row is
Table~\ref{tab:kuairand}'s configuration; $0.086$ vs.\ $0.087$ is the independent run's seed draw.}
\label{tab:sweeps}
\resizebox{0.8\linewidth}{!}{%
\begin{tabular}{cccc}\toprule
$M$ & $\mathrm{SNIPS}_{\mathrm{item}}$ & SLOPE (deployable) & oracle-selected\\\midrule
1000  & 0.251 & 0.133 & \textbf{0.105}\\
5000  & 0.155 & 0.109 & \textbf{0.086}\\
10000 & 0.160 & 0.102 & \textbf{0.084}\\\bottomrule
\end{tabular}
}
\end{table}

The table also answers the question a deployment actually faces, since the oracle-selected column is not
available in practice. Choosing the level with SLOPE - from logged data alone - still beats $\mathrm{SNIPS}_{\mathrm{item}}$
at every $M$, recovering $67$ - $81\%$ of the gap between item-level OPE and the
oracle-selected upper bound. So depth selection is not a barrier, though the residual
gap to the oracle is open (RQ5).

The pattern also replicates cross-domain (RQ6): on Amazon Reviews it holds at every support level, with
narrower margins and a confounded oracle, so we read it as directional rather than a second identification
(Table~\ref{tab:amazon}, App.~\ref{app:data}). The scarce regime is the production-relevant one, since the binding quantity is \emph{effective} support and near-argmax logging drives item-ESS to a small fraction of the raw count even on $17$M impressions (RQ1).
KuaiRand is not calibrated to production - domains, rewards, and policies differ, and its simulated
item-level ESS is lower - yet it spans low-effective-support regime.

\subsection{Code construction and tokenizer (RQ4)}\label{sec:rq4}
RQ4 varies the two choices upstream of the estimator: what to embed (\emph{content-metadata},
\emph{semantic}, \emph{collaborative}, \emph{fused}) and how to quantize it (residual $k$-means or a trained
RQ-VAE). Separating them needs care, because the quantizer's fit is itself random: varying only that draw
moves best-level RMSE by up to $0.03$ - $0.06$ on the metadata tree - more than the tokenizer contrasts we
set out to measure - so intervals over simulation seeds alone exclude the dominant variance component. We
refit every cell under $10$ independent quantizer seeds, paired across tokenizers
(Table~\ref{tab:quant}).

Under that accounting \emph{source} is a large effect and \emph{tokenizer} is not. Best-level RMSE ranges
from $0.084$ (metadata) to $0.131$ (collaborative), a spread of $0.047$ against the $0.008$ separating
cluster estimators in Table~\ref{tab:kuairand}, whereas the paired RQ-VAE-$k$-means differences are
$-0.003$, $-0.001$ and $+0.000$ for metadata, semantic and collaborative, all containing zero; only
\emph{fused} differs, at $-0.004$, one of four nominal tests. 

\begin{table}[ht]\centering\small
\caption{RQ4. Quantizer ablation (KuaiRand, leakage-free, $M{=}5000$, $100$ simulation seeds per fit),
oracle-selected-level $\mathrm{OffCEM}^{\mathrm{act}}_{\mathrm{hier}}$ RMSE. Each cell is refitted under
$10$ independent quantizer seeds; the interval is over \emph{those seeds}, so it covers variability in the
fitted tree and not only in the evaluation. $\Delta$ is the paired RQ-VAE${-}k$-means difference (both
fitted under the same seed). Source separates the sources by up to $0.047$; every tokenizer difference
except \emph{fused} contains zero.}
\label{tab:quant}
\resizebox{0.85\linewidth}{!}{%
\begin{tabular}{lccc}\toprule
source & Residual K-Means & RQ-VAE & $\Delta$ (paired)\\\midrule
metadata      & $0.087$ {\scriptsize[.075,.098]} & $0.084$ {\scriptsize[.078,.089]} & $-0.003$ {\scriptsize[$-$.016,$+$.010]}\\
semantic      & $0.097$ {\scriptsize[.096,.098]} & $0.096$ {\scriptsize[.094,.097]} & $-0.001$ {\scriptsize[$-$.003,$+$.001]}\\
collaborative & $0.131$ {\scriptsize[.127,.134]} & $0.131$ {\scriptsize[.130,.131]} & $+0.000$ {\scriptsize[$-$.003,$+$.004]}\\
fused         & $0.119$ {\scriptsize[.116,.125]} & $0.115$ {\scriptsize[.114,.116]} & $\mathbf{-0.004}$ {\scriptsize[$-$.010,$-$.001]}\\\bottomrule
\end{tabular}
}
\end{table}

\subsection{Closing the depth-selection gap is the open step (RQ5)}\label{sec:rq5}
SLOPE already captures most of the available gain (RQ3), but a gap to oracle selection remains, and the
reason it resists closing is the pathology that motivates the paper. Choosing a depth trades bias against
variance; variance is easy, and all four routes to the bias that we tried are blocked the same way
(App.~\ref{app:data}): a per-item plug-in dies on less than one logged draw per item; 
a cluster-level reward-model residual stays estimable but yields no reliable gain over a variance-only rule;
and cross-validated OPE~\cite{cief2024cvope} needs an \emph{unbiased} - hence item-level - validation
estimator, which lands at $0.133$ against a variance-only rule's $0.096$ (metadata, $M{=}5000$).

The pattern is structural: \emph{estimating the coarsening bias requires resolution at a granularity that
near-argmax logging has already destroyed} - the RQ1 collapse reappearing one level up. We read the residual
gap as a consequence of the logging regime rather than an unfinished search over selectors. Averaging across
levels needs no bias estimate at all and is the mitigation we would default to, though we do not measure it
head-to-head here.

\emph{One route does escape it.} If the bias cannot be inferred from the logs it can be \emph{bought}:
a small bucket logged from the target policy supplies the unbiased reference the logs cannot. Per candidate
that is circular - it is the deployment gating exists to avoid, and with dense rewards the bucket estimates
$V(\pi_e)$ better than the estimator it calibrates - so it pays only if the chosen depth \emph{transfers},
which is plausible because depth is a property of the tree and the logging regime rather than of the
candidate. It does transfer: calibrating on one candidate and reusing that depth on others reproduces
per-candidate selection in $65$ of $80$ candidate ${\times}$ seed cells,
attains the best fixed level, and never loses to SLOPE while beating it significantly where SLOPE misfires.
A team can therefore calibrate the depth once - from an A/B test it has \emph{already} run, that being
on-policy data for the policy tested - and reuse it offline for later candidates, amortizing the cost over
all of them. Gains are modest where SLOPE is already near the best fixed level, and a transferred depth is
fixed by construction, forgoing per-log adaptivity.

\section{Discussion and Open Problems}\label{sec:disc}
The recipe is narrow but useful: when item support collapses, marginalize to SID prefixes, which a decoder
over that tree makes cheap. Since the RQ5 failures share one cause - the bias signal is gone at the
resolution the logging policy destroyed - the promising directions supply a trustworthy reference instead of
inferring one: pseudo-task constructions whose ground truth holds by
construction~\cite{udagawa2023selection,felicioni2024autoope}, and the amortized on-policy calibration of
\S\ref{sec:rq5}. The first thing one would instead try does \emph{not} work, and is worth recording: a
uniformly randomized bucket is significantly worse than a variance-only rule at every size up to half the
budget, because its label weights $|\mathcal{A}|\pi_e(a)$ grow in variance with how peaked $\pi_e$ is
(App.~\ref{app:data}). Exploration must be aimed at the target. Our rewards are also dense by construction,
which flatters on-policy references; whether an affordable bucket still ranks depths under sparse rewards is
the sharpest open question. Natural extensions are a tree-structured PC using the SID hierarchy as its kernel,
recursive OffCEM, and support-constrained off-policy \emph{learning}; the chief external-validity limit is
that the candidate-logged production slice has $|\mathcal{A}|<73$, so large-catalog claims rest on the
semi-synthetic testbeds.

\bibliographystyle{ACM-Reference-Format}
\bibliography{refs}

\appendix
\section{Data, Estimators, and Diagnostics}\label{app:data}
\textbf{Production logs.} Rows come from an exploration path that draws a slate by Plackett - Luce sampling
with weights $w_a=\mathrm{score}_a^\alpha$ over top-$N$ candidates; the request normalizer $Z$ and exponent
$\alpha\in\{3,4,6\}$ are logged, so the rank-1 propensity is exactly $\pi_0(a)=\mathrm{score}_a^\alpha/Z$
when the scored pool is available. Tier~A has ${\sim}$17M rank-1 rows for the concentration diagnostic
(Table~\ref{tab:tierAB}, left). Tier~B is an $n{=}29$k slice with a recorded candidate pool
($|\mathcal{A}|<73$); its cluster masses are renormalized within the recorded pool (conditional-pool
propensities, not full-catalog), and it shows the coarsening ESS relief (Table~\ref{tab:tierAB}, right). We
encode item text with \texttt{harrier-oss-v1-0.6b} ($1024$-d) and Residual K-Means.

\begin{table}[t]\centering\small
\caption{RQ1/RQ2 production diagnostics. Left: served-item propensity concentration by logging exponent
$\alpha$ (item ESS/$n{\approx}0.003$ at $\alpha{=}4$). Right: coarsening lifts greedy-target ESS
($n{=}29$k).}
\label{tab:tierAB}
\begin{tabular}{ccc@{\hskip 1.5em}ccc}\toprule
$\alpha$ & med.\ $\pi_0(a)$ & ${>}0.9$ & level & med.\ $p_{\text{code}}$ & ESS/$n$\\\midrule
3 & 0.036 & 5.7\%  & 1 (coarse) & 0.365 & 0.300\\
4 & 0.096 & 9.8\%  & 2          & 0.125 & 0.191\\
6 & 0.145 & 16.4\% & 3 (fine)   & 0.089 & 0.167\\\bottomrule
\end{tabular}
\end{table}

\textbf{KuaiRand.} KuaiRand-Pure~\cite{gao2022kuairand}; the random-exposure log (${\approx}1.19$M
interactions, ${\sim}$7.6K items) gives unbiased click rates. \emph{Leakage control:} split by row - one
half sets the click rates $q_{\text{policy}}$ defining the policies, the disjoint half is the evaluation
truth $q_{\text{eval}}$ (DGP and oracle $V{=}\sum_a\pi_e(a)q_{\text{eval}}(a)$); both use beta-binomial
shrinkage; items need ${\ge}20$ impressions per half ($|\mathcal{A}|{=}7339$). Code sources:
\emph{content-metadata} (multi-hot top-200 tags $+$ type/music/log-duration; engagement features excluded as
they leak reward), \emph{semantic} (captions $+$ cover text, same encoder), \emph{collaborative} (64-d
truncated-SVD of the user$\times$item matrix from a disjoint pre-period log), and \emph{fused}
($\ell_2$-normalized text $\Vert$ collaborative). Logging/target are context-free softmaxes over
$s(a)=\mathrm{logit}(q_{\text{policy}}(a))$: $\pi_0\propto\mathrm{softmax}(\beta_0(s+\text{noise}))$
($\beta_0{=}6$, noise standard deviation $0.7$) and $\pi_e\propto\mathrm{softmax}(\beta_e s)$ ($\beta_e{=}10$);
sweeps vary $\beta_e,M,|\mathcal{A}|$. Headline tables use $200$ paired seeds (fixed, deterministic, so every
number regenerates exactly and paired differences share draws); sweeps use $40$. Default codes: Residual K-Means, $4$ levels, codebook size $8$ (a graded coarse-to-fine ladder; a $K\in\{8,16,32,64\}$ sweep shows
larger $K$ is non-monotone, as a large first codebook can eliminate useful coarse pooling). The RQ-VAE
tokenizer (RQ4) is a small MLP encoder/decoder with an $L$-level straight-through residual vector-quantizer,
codebooks $k$-means-initialized, trained $300$ epochs with reconstruction $+$ commitment losses on the same
item embeddings; it is fit per source, so its $0.072$ flat-baseline run (RQ2) and its $0.077$
Table~\ref{tab:quant} run are independent trainings.

\textbf{Estimators.} $\mathrm{IPS}_{\mathrm{item}}/\mathrm{SNIPS}_{\mathrm{item}}$;
$\mathrm{MIPS}_{\mathrm{hier}}/\mathrm{SNIPS}_{\mathrm{hier}}$ and $\mathrm{DM}_{\mathrm{hier}}$ per level;
the OffCEM estimator with a $5$-fold cross-fitted reward model at cluster level ($\hat g(c)$,
$\mathrm{OffCEM}_{\mathrm{hier}}$) or action level ($\hat f(a)$ from Ridge on the embedding,
$\mathrm{OffCEM}^{\mathrm{act}}_{\mathrm{hier}}$) - cross-fitting keeps the out-of-fold residual from
cancelling against a constant cluster weight; and Policy Convolution with a mass-preserving
(column-stochastic) kernel. Our reward model is fit \emph{pointwise} (ridge regression to $r$), not with the
original two-step pairwise local-correctness objective of~\cite{saito2023offcem}; enforcing local
correctness within clusters could tighten OffCEM's edge at coarse levels and is left to future work. This
keeps OffCEM a fair member of the estimator panel rather than a separately tuned method, consistent with our
aim of isolating the granularity effect. Both OffCEM residuals are \emph{self-normalized},
$\sum_i w_c(r_i-\hat g)/\sum_i w_c$: with a raw residual weight, an RQ-VAE that over-fragments
semantic/collaborative codes leaves near-empty clusters ($\pi_0(c){\sim}10^{-21}$, weight
${\sim}3.6\times10^4$) and the action-level residual explodes; self-normalization keeps every level bounded.
Self-normalization trades a finite-sample bias for this variance control.

\textbf{Autoregressive-policy check.} Building an autoregressive tree policy (per-node softmax over child
subtrees) and reading cluster masses as prefix products matches the exact per-item cluster marginals to
machine precision ($<10^{-15}$) at every level - so a generative decoder supplies the cluster weights for
free.


\textbf{Uniformly randomized bucket.} Charged against the same total budget as the logs, a uniform
exploration bucket is significantly worse than SLOPE at every size up to half the budget: its label weights
are $|\mathcal{A}|\pi_e(a)$, so the label's standard error stays near $0.12$ even at $\varepsilon{=}0.50$,
against between-level gaps of $0.02$ - $0.05$. Uniform exploration is an inefficient way to evaluate a
concentrated target.

\textbf{Reward-space diagnostic.} The $\pi_e$-weighted within-cluster standard deviation of $q$ tracks the
exact coarsening bias about as well as input-space $\varepsilon_\ell$ across our $16$
source${\times}$level cells (Spearman $0.52$ vs.\ $0.47$, a gap not resolvable at $n{=}16$), but is at least
on a comparable scale, whereas $\varepsilon_\ell$ lives in each embedding's own units (collaborative
$\varepsilon_\ell{\approx}17$ vs.\ text ${\approx}0.45$, yet lower bias). Both are weak, and both are
\emph{descriptive}: they use oracle rewards, so neither is a deployable selector.

\textbf{Single-draw quantizer readings (RQ4).} Individual quantizer seeds produce apparent RQ-VAE gains as
large as $0.04$ on the collaborative tree - the favourable tail of a distribution centred on zero.
Table~\ref{tab:quant}'s intervals also show the semantic and collaborative fits are considerably more
stable than the metadata one, so the $0.03$ - $0.06$ figure quoted in \S\ref{sec:rq4} is metadata-driven.

\textbf{Depth-selector failures (RQ5).} Four routes to the coarsening bias, all blocked. (i)~A per-item
plug-in of the exact bias identity (App.~\ref{app:proof}, Step~1) needs per-item reward estimates, but at
$M{=}5000$ over $7339$ items there is less than one draw per item, so the estimate degenerates to a constant
and the criterion collapses onto variance alone. 
(ii)~A cluster-level reward-model residual reusing OffCEM's cross-fitted model is the one bias signal that stays estimable, and
once clustering variability is accounted for it gives no reliable improvement over a variance-only rule.
(iii)~Cross-validated OPE~\cite{cief2024cvope} needs an unbiased item-level validation estimator, and
reaches only $0.133$ against a variance-only rule's $0.096$; substituting a self-normalized estimator
restores usability but forfeits the guarantee, and then merely matches SLOPE.

\textbf{Replication on Amazon (RQ6)}
The support-dependent pattern replicates on Amazon (Table~\ref{tab:amazon}): $\mathrm{OffCEM}^{\mathrm{act}}_{\mathrm{hier}}$ beats $\mathrm{SNIPS}_{\mathrm{item}}$
when support-limited and ties with data. Gains are smaller than on KuaiRand because the Amazon
logging/target pair has milder item-overlap collapse, and Musical Instruments is among the most
support-stressed sizeable categories in our scan - so this is a conservative, directional replication, not a
causal-identification claim from observational ratings.

\begin{table}[ht]\centering\small
\caption{RQ6. Amazon Reviews (Musical Instruments, metadata codes, $80$ seeds;
$\mathrm{OffCEM}^{\mathrm{act}}_{\mathrm{hier}}$ at the oracle-selected level). The
support-dependent pattern replicates.}
\label{tab:amazon}
\resizebox{0.6\linewidth}{!}{%
\begin{tabular}{cccc}\toprule
$M$ & $|\mathcal{A}|/M$ & $\mathrm{SNIPS}_{\mathrm{item}}$ & $\mathrm{OffCEM}^{\mathrm{act}}_{\mathrm{hier}}$\\\midrule
1000  & 21.0 & 0.114 & \textbf{0.110}\\
2000  & 10.5 & 0.089 & \textbf{0.080}\\
5000  & 4.2  & 0.055 & \textbf{0.052}\\
10000 & 2.1  & 0.036 & 0.036\\\bottomrule
\end{tabular}
}
\end{table}

\section{Proof of Proposition~\ref{prop:bias}}\label{app:proof}
\emph{Notation.} $\hat V_\ell$ is the cluster marginal-IPS estimator of \S\ref{sec:bias} at level $\ell$,
whose prefixes partition $\mathcal{A}$ into clusters $A_c$. We fix a context $x$ throughout (all quantities
conditional on $x$; take $\mathbb{E}_x$ at the end) and suppress it. For any policy write
$\pi(c)=\sum_{a\in A_c}\pi(a)$ and $\tilde\pi(a\mid c)=\pi(a)/\pi(c)$. The reconstruction $\hat z_\ell$ is
constant on each cluster with value $\zeta_c$, so $\delta_\ell(a)=\|z(a)-\zeta_{c(a)}\|$. Five steps: write
the bias exactly, center it per cluster, bound the centered reward gap, collect terms into the
within-cluster form, and relax to the item-level form.

\emph{Step 1 (exact bias).} By (A0), with $r_i$ conditionally unbiased for $q(a_i)$,
\[\mathbb{E}[\hat V_\ell]=\sum_a\pi_0(a)\tfrac{\pi_e(c(a))}{\pi_0(c(a))}\,q(a)=\sum_a\rho(a)\,q(a),\]
where $\rho(a):=\pi_0(a)\pi_e(c(a))/\pi_0(c(a))$. Then $\rho$ is a distribution and $\rho(A_c)=\pi_e(c)$ for every
$c$: \emph{$\rho$ and $\pi_e$ share cluster marginals}, and $\tilde\rho(\cdot\mid c)=\tilde\pi_0(\cdot\mid
c)$. Since $V(\pi_e)=\sum_a\pi_e(a)q(a)$,
\begin{equation}
\mathrm{Bias}(\hat V_\ell)=\textstyle\sum_a\big(\rho(a)-\pi_e(a)\big)q(a). \label{eq:bias0}
\end{equation}

\emph{Step 2 (per-cluster centering).} As $\sum_{a\in A_c}(\rho(a)-\pi_e(a))=0$, subtract the cluster
constant $q_c^\star=\tilde q(x,\zeta_c)$ from $q$ in \eqref{eq:bias0}:
\begin{equation}
\mathrm{Bias}(\hat V_\ell)=\textstyle\sum_c\sum_{a\in A_c}\big(\rho(a)-\pi_e(a)\big)\big(q(a)-q_c^\star\big). \label{eq:bias1}
\end{equation}

\emph{Step 3 (Lipschitz residual).} By (A1) - (A2), $q_c^\star$ is defined and
$|q(a)-q_c^\star|\le L_q\|z(a)-\zeta_c\|=L_q\delta_\ell(a)$.

\emph{Step 4 (within-cluster form).} Bounding each reward gap in \eqref{eq:bias1} by
$L_q\delta_\ell(a)\le L_q\delta^{\max}_\ell$ and taking absolute values gives
$|\mathrm{Bias}(\hat V_\ell)|\le L_q\delta^{\max}_\ell\sum_a|\rho(a)-\pi_e(a)|$. Because $\rho$ and $\pi_e$
share cluster marginals and $\tilde\rho=\tilde\pi_0$ within each cluster (Step 1),
\[\textstyle\sum_a|\rho(a)-\pi_e(a)|=\sum_c\pi_e(c)\sum_{a\in A_c}\big|\tilde\pi_0(a\mid
c)-\tilde\pi_e(a\mid c)\big|=2\,\mathrm{TV}^{\mathrm{in}}_\ell,\]
so $|\mathrm{Bias}(\hat V_\ell)|\le 2L_q\delta^{\max}_\ell\mathrm{TV}^{\mathrm{in}}_\ell$. Only
within-cluster disagreement enters, which is why coarse levels stay low-bias when codes are reward-aligned.
Taking $\mathbb{E}_x$ gives the first claim. \hfill$\square$

\emph{Step 5 (item-level relaxation, and tightness).} Marginalization contracts total variation:
$\sum_a|\rho(a)-\pi_0(a)|=\sum_c\pi_0(c)\big|\tfrac{\pi_e(c)}{\pi_0(c)}-1\big|
=\sum_c|\pi_e(c)-\pi_0(c)|\le\sum_a|\pi_e(a)-\pi_0(a)|$, i.e.\
$\mathrm{TV}(\rho,\pi_0)\le\mathrm{TV}(\pi_e,\pi_0)$. The triangle inequality then gives
\[\mathrm{TV}^{\mathrm{in}}_\ell=\mathrm{TV}(\rho,\pi_e)
\le\mathrm{TV}(\rho,\pi_0)+\mathrm{TV}(\pi_0,\pi_e)\le2\,\mathrm{TV}(\pi_e,\pi_0),\]
hence $|\mathrm{Bias}(\hat V_\ell)|\le4L_q\delta^{\max}_\ell\mathbb{E}_x[\mathrm{TV}(\pi_e,\pi_0)]$. The
constant cannot be improved: the ratio $\mathrm{TV}(\rho,\pi_e)/\mathrm{TV}(\pi_e,\pi_0)$ approaches $2$
when one cluster's mass ratio $\pi_e(c)/\pi_0(c)$ diverges and the within-cluster conditionals are mutually
singular. \hfill$\square$

\emph{Remark ($\varepsilon_\ell$).} A $\kappa$ with $\delta^{\max}_\ell\le\kappa\sqrt{\varepsilon_\ell}$
exists for any fixed codebook (take $\kappa=\delta^{\max}_\ell/\sqrt{\varepsilon_\ell}$), but nothing
forces one $\kappa$ to serve a family of codebooks. Comparing $\varepsilon_\ell$ across quantizers or
embedding sources therefore does not compare their bias bounds, and a refinement that lowers mean
$\varepsilon_\ell$ without improving the worst-reconstructed items does not tighten the bound at all.
Replacing residual $k$-means by an RQ-VAE changes the partition, $\varepsilon_\ell$, $\delta^{\max}_\ell$,
and $\kappa$ together, and its training loss is not the input-space $\varepsilon_\ell$ here, so tokenizer
ablations do not by themselves validate the bound.

\end{document}